\documentclass{article}

\usepackage[preprint,nonatbib]{neurips_2020}

\usepackage[utf8]{inputenc} 
\usepackage[T1]{fontenc}    
\usepackage{hyperref}       
\usepackage{url}            
\usepackage{booktabs}       
\usepackage{amsfonts}       
\usepackage{nicefrac}       
\usepackage{microtype}      
\usepackage{bm}
\usepackage{amsthm}
\usepackage[makeroom]{cancel}
\usepackage[shortlabels]{enumitem}

\usepackage{graphicx}
\usepackage{amsmath}
\usepackage{wrapfig}
\usepackage{subcaption}
\usepackage{diagbox}
\usepackage{multirow}

\usepackage{algorithm} 
\usepackage{algorithmic}

\renewcommand{\epsilon}{\varepsilon}

\newcommand{\cD}{\mathcal{D}}

\newcommand{\cL}{\mathcal{L}}

\newcommand{\cS}{\mathcal{S}}

\newcommand{\cX}{\mathcal{X}}

\newcommand{\cZ}{\mathcal{Z}}

\newcommand{\bbE}{\mathbb{E}}

\newcommand{\bbR}{\mathbb{R}}

\newcommand{\bzero}{\mathbf{0}}

\DeclareMathOperator*{\argmax}{argmax}

\usepackage{xcolor}
\usepackage{float}

\definecolor{asparagus}{rgb}{0.53, 0.66, 0.42}
\definecolor{darkgreen}{rgb}{0.15, 0.6, 0.3}
\newtheorem{prop}{Proposition}
\usepackage{wrapfig}
\definecolor{darkgreen}{rgb}{0.0, 0.5, 0.0}
\title{Decoder-free Robustness Disentanglement\\
 without (Additional) Supervision}

 \author{%
 Yifei Wang\thanks{Work was done during an internship at Huawei Noah's Ark Lab.} \\
 Peking University\\
 \texttt{yifei\_wang@pku.edu.cn} \\
 \And
Dan Peng \\
Huawei Noah's Ark Lab \\
\texttt{pengdancs@gmail.com} \\
\And
Furui Liu \\
Huawei Noah's Ark Lab \\
\texttt{liufurui2@huawei.com} \\
\And
Zhenguo Li \\
Huawei Noah's Ark Lab \\
\texttt{li.zhenguo@huawei.com} \\
\And
Zhitang Chen \\
Huawei Noah's Ark Lab \\
\texttt{chenzhitang2@huawei.com} \\
\And 
Jiansheng Yang \\
Peking University\\
\texttt{yjs@math.pku.edu.cn} \\
}

\begin{document}
\maketitle
\begin{abstract}
Adversarial Training (AT) is proposed to alleviate the adversarial vulnerability of machine learning models by extracting only robust features from the input, which, however, inevitably leads to severe accuracy reduction as it discards the non-robust yet useful features. This motivates us to preserve both robust and non-robust features and separate them with disentangled representation learning. Our proposed Adversarial Asymmetric Training (AAT) algorithm can reliably disentangle robust and non-robust representations without additional supervision on robustness. Empirical results show our method does not only successfully preserve accuracy by combining two representations, but also achieve much better disentanglement than previous work. 
\end{abstract}

\section{Introduction}
A well-known obstacle of machine learning is the existence of adversarial examples. A small invisible perturbation to the input can lead to dramatic misbehavior of neural networks \cite{szegedy2013intriguing}, raising huge concern about the vulnerability of machine learning. Various attack and defense algorithms have been developed ever since, like a cat and mouse game \cite{chakraborty2018adversarial}.

Adversarial Training \cite{goodfellow2014explaining} is proposed to train a classifier with adversarial examples and effectively makes it more robust to perturbations. Perhaps surprisingly, it is found that the robust features, i.e., image features utilized by robust classifiers, are perceptually aligned with humans \cite{schmidt2018adversarially}. On the contrary, the non-robust features from a standard classifier, are also useful for classification but look like plain noise to humans. This indicates that adversarial training is founded on a human-centric perspective \cite{ilyas2019adversarial}, that  it enforces neural networks to achieve the robustness defined by human recognition, i.e., robustness against perturbations, which, however, may conflict with the nature of neural networks. As a result, adversarial training of neural networks will inevitably lead to severe accuracy reduction in the classification of natural images \cite{tsipras2018robustness}.

Nevertheless, robustness is desirable in some scenarios where humans are involved in the loop. In the meantime, non-robust features also matter for accuracy, and it seems unwise to discard them as in adversarial training.
As Bengio et al. \cite{bengio2013representation} put it, \emph{the most robust approach to feature learning is to disentangle as many factors as possible, discarding as little information about the data as is practical}.
Motivated by this, instead of keeping either of them, we propose to map robust and non-robust features into two disentangled representations. Thus, both robust and non-robust features are not only preserved but also well separated. Afterwards, we can obtain a robust or non-robust classifier with only one of the representations, or achieve better  accuracy by combining two representations when necessary.

Learning deep representations where different semantic aspects of data are structurally disentangled is of central importance for training robust models \cite{bengio2013representation,suter2018robustly}. To achieve disentanglement, supervised methods require access to additional supervision, in the form of pairwise data sharing the same attributes \cite{mathieu2016disentangling}, or the ground truth generative mechanism \cite{kulkarni2015deep}, etc. But those supervisions are hardly available in practice. Alternatively, some focus on disentangling latent factors purely from unsupervised data \cite{higgins2017beta,kim2018disentangling}, which, however, are  challenged lately \cite{locatello2018challenging} as their disentanglement scores are heavily influenced by randomness, and the disentanglement shows no clear benefit for downstream tasks.

Our method has advantages over both diagrams. On the one hand, we can disentangle robust and non-robust features without additional supervision on robustness. With class-labeled data, previous work \cite{siddharth2017learning,gowal2019achieving} can only disentangle w.r.t. class itself, while ours can disentangle w.r.t. robustness, which is not directly given by data. On the other hand, the unsupervised methods mostly rely on Variational Autoencoders \cite{kingma2014auto} with unstable performance \cite{locatello2018challenging}, while our model is more efficient and effective as it is deterministic, decoder-free, and able to produce successful disentanglement with little random variability. Last but not least, our disentanglement shows clear benefits for downstream applications, such as standard and adversarial predictions, as well as adversarial detection and calibration. 

But how to achieve the disentanglement remains unclear. For a natural image, its robust and non-robust features are entangled together, and we hardly know the ground truth of either of them. Ilyas et al. \cite{ilyas2019adversarial} developed an iterative optimization scheme that constructs images with only robust or non-robust features of a natural image, for which we call \emph{pseudo-inputs}. Experiments show that they can achieve a limited degree of disentanglement, but cause even worse accuracy reduction because the generation of pseudo-inputs leads to a great loss of details in the raw images. 

Our disentanglement method is based on the idea of \emph{pseudo-pairs} instead. We notice that essentially the process of adversarial attack is about modifying the non-robust features such that they belong to a wrong class and lead to misclassification of the images. Therefore, a misclassified adversarial example is supposed to contain robust and non-robust features about different classes, and the combination of a natural and an adversarial example yields a \emph{pseudo-pair} for robustness disentanglement. Based on this insight, we propose \emph{Adversarial Asymmetric Training (AAT)} that assigns asymmetric labels to robust and non-robust representations, and the asymmetry disentangles them apart. Vanilla Adversarial Training extracts robust features alone and fails at preserving standard accuracy, while our AAT extracts both kinds of features with disentanglement, and makes it possible to preserve accuracy by combining two representations. Compared to the pseudo-input method \cite{ilyas2019adversarial}, the disentanglement with pseudo-pairs preserves the details of the images and achieves much better accuracy and disentanglement. Besides, our method trains models end-to-end with much less computation.

\section{Method}

\subsection{Notations and Preliminary}
\label{sec:preliminary}
\textbf{Standard training.}
Consider image classification with labeled training data $\cD_{train}=\{(x,y)\}$, where $x\in\bbR^D$ is a $D$-dimensional input image, $y\in\{1,\dots,C\}$ denotes its class label, and $C$ is the number of classes. We can train a classifier $h$ with parameters $\theta$ by minimizing training loss as 
\begin{equation}
\min_\theta\bbE_{(x,y)\sim\cD_{train}}\cL(\theta,x,y),\quad\quad\cL(\theta,x,y)=l(h(x;\theta),y),  \label{eqn:standard-training}
\end{equation}
where $l(\cdot, \cdot)$ denotes the loss function, e.g. cross entropy, and $h(x;\theta)$ is the predicted probability distribution over $C$ classes. Assume the classifier has good \text{standard accuracy} after training. 

\textbf{Adversarial example.} However, the standard classifier  can be easily fooled by adversarial examples, generated with small perturbation $\delta$ to the input image that maximizes the loss function \cite{szegedy2013intriguing}, 
\begin{eqnarray}
x^s_{adv}=x+\underset{\delta\in\cS}{\arg \max }\,\, \cL(\theta,x+\delta, y),\label{eqn:untargeted-attack-objective}
\end{eqnarray}
where $\cS=\{\delta\,|\,\left\|\delta\right\|_p \leq \varepsilon\}$ is the set of all feasible perturbations within $\ell_p$ norm constraint. We refer to the classification accuracy under attack as \text{robust accuracy}.

\textbf{Adversarial training.} To alleviate adversarial attack, \textit{Adversarial Training} (AT) \cite{goodfellow2014explaining} is proposed to train a robust classifier by solving the following robust optimization problem \cite{madry2018towards}
\begin{equation}
\min _{\theta} \mathbb{E}_{(x, y) \sim\cD_{train}}\left[\max _{\delta \in \mathcal{S}} \cL(\theta, x+\delta, y)\right]. \label{eqn:adversarial-training}
\end{equation}
Specifically, for a data sample $(x, y)$, we first solve the inner loop and get the adversarial example $x_{adv}$, and then update parameters $\theta$ with adversarial pair $(x_{adv},y)$ for the outer loop.

\textbf{Robust and non-robust features.} 
Define a feature as a function mapping from the input space $\cX$ to real numbers $h:\cX\to\bbR$. With a specified data distribution $\cD$ and an adversarial configuration $\cS$, we give formal definitions of robust and non-robust features for binary classification ($C=2$).
\begin{itemize}
    \item 
    We call a feature $h$ \emph{$\rho$-useful} ($\rho$ > 0) if it is correlated with the true label in expectation, i.e.,
$        \mathbb{E}_{(x, y) \sim \mathcal{D}}[y \cdot h(x)] \geq \rho.$
    \item 
    Suppose we have a \(\rho\)-useful feature $h$, we refer to \(h\) as a \emph{robust feature} (formally a $\gamma$-robustly useful feature) if \(h\) remains \(\gamma\)-useful $(\gamma>0)$ under adversarial perturbation, i.e.,
    $ \mathbb{E}_{(x, y) \sim \mathcal{D}}\left[\min _{\delta \in\cS} y \cdot h(x+\delta)\right] \geq \gamma.$
    \item We refer to \(h\) as a \emph{non-robust feature} (formally a $\gamma$-non-robustly useful feature) 
     if it is $\rho$-useful for some $\rho>0$, but not $\gamma$-robust ($\gamma> 0$).
\end{itemize}
In other words, both robust and non-robust features are useful for classification, and they differ merely in their behaviors under adversarial attack. As for their disentanglement, we should encourage both of them to attain better usefulness (higher $\rho$), while encourage robust features to attain better robustness (higher $\gamma$) and encourage non-robust features to attain better non-robustness (lower $\gamma$). We also give formal definitions for robust and non-robust representations likewise in Appendix \ref{sec:representation-definitions} and a discussion of the accuracy-robustness dilemma in Appendix \ref{sec:dilemma}.

\subsection{Model}
\label{sec:model}

\begin{figure}[t]
\centering
\includegraphics[width=\linewidth]{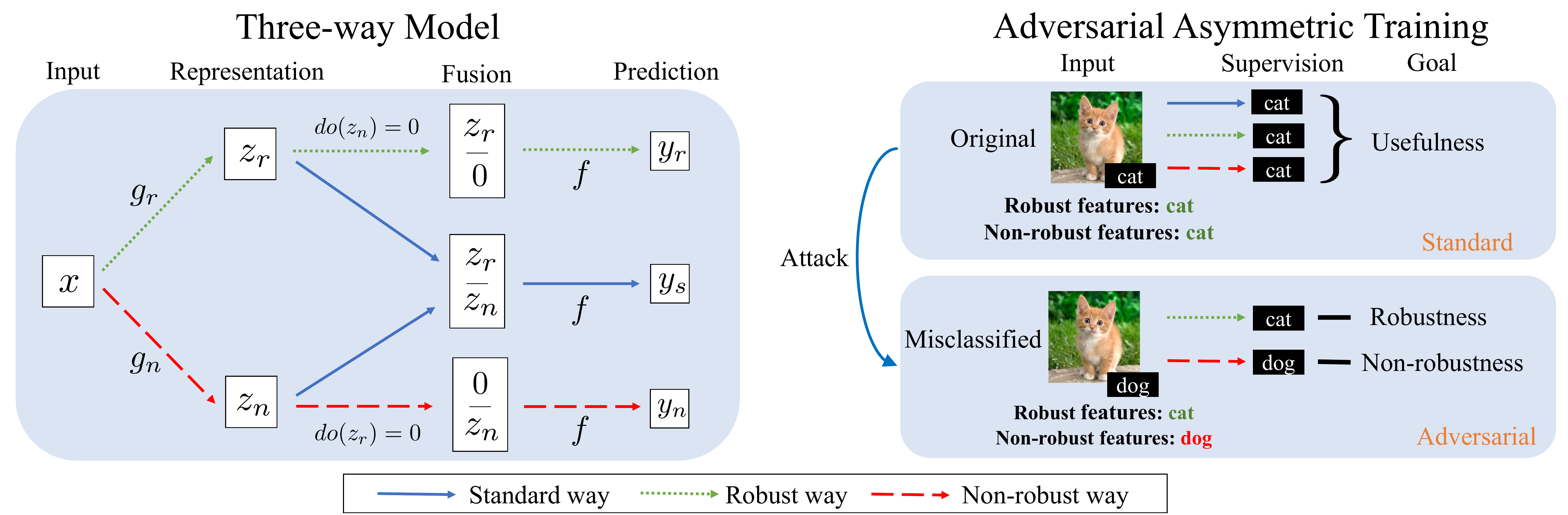}
\caption{Left: a diagram of our proposed three-way model. 
We encode an input image $x$ into the robust and non-robust representations $z_r$ and $z_n$ with two encoders $g_r$ and $g_n$. Afterwards, we use three ways to combine the two representations and predict the labels with a shared classifier $f$. Right: an illustration of our proposed Adversarial Asymmetric Training with an image of a cat that is misclassified as a dog after adversarial perturbation. We assign asymmetric supervisions to the two representations to achieve robustness disentanglement.}
\label{fig:diagram}
\end{figure}

We start by proposing a discriminative model for disentanglement. Given an input image $x$, we extract two representations through two encoders $g_r$ and $g_n$ with parameters $\theta_r$ and $\theta_n$,
\begin{subequations}
\label{eqn:representations}
\begin{align}
\text{robust representation:~}  & z_{r}=g_r(x;\theta_r);\label{eqn:robust-representation}\\
\text{non-robust representation:~} & z_{n}=g_{n}(x;\theta_{n}).\label{eqn:non-robust-representation}
\end{align}
\end{subequations}
and they are supposed to extract robust and non-robust features of the image, respectively, and thus disentangle them apart.  
Afterwards, the two representations are combined in three different ways and get predictions with a shared classifier $f$ on the top,
\begin{subequations}
\label{eqn:three-ways}
\begin{flalign}
\text{standard way:~}& h_s(x;\theta) = f([z_r,z_{n}]); \label{eqn:standard-way}\\
\text{robust way:~} & h_r(x;\theta) = f([z_r, \bzero]); \label{eqn:robust-way}\\
\text{non-robust way:~} & h_{n}(x;\theta) = f([\bzero, z_{n}]);\label{eqn:non-robust-way}
\end{flalign}
\end{subequations}
where we denote $[\cdot,\cdot]$ for vector concatenation, and $\bzero$ for a zero vector of equal size as $z_r$ and $z_n$. See Figure \ref{fig:diagram} (left) for illustration. In $h_r$ and  $h_n$, we set one of the representations to a constant vector to make it non-informative. From a causal view \cite{zhang2020causal}, essentially we are performing interventions \cite{pearl2009causality} to eliminate the robust or non-robust features, i.e., $do(z_r)=0$ or $do(z_n)=0$. 

\subsection{Learning}
\label{sec:learning}
In this part, we describe two training objectives for our model to achieve the disentanglement.

\subsubsection{Standard Training}
\label{sec:standard-training}
By definition both robust and non-robust features are useful for prediction, thus all three ways should have high standard accuracy. Therefore, given a natural data pair $(x,y)$, 
we assign a standard classification loss to each of the three ways $h_s, h_r, h_n$, and get the standard training loss as
\begin{equation}
    \begin{aligned}
        \cL_{ST}(\theta,x,y)=\,& {\color{blue} l(h_s(x;\theta),y)} + 
        {\color{darkgreen} l(h_r(x;\theta),y)}  
        +{\color{red} l(h_{n}(x;\theta),y)}.        
    \end{aligned}
    \label{eqn:total-standard-training-loss}
\end{equation}
As a result, both representations are learned to be \text{useful} for prediction. Nevertheless, we can see that the supervisions for robust and non-robust representations are totally \textbf{symmetric} in the standard training loss. Therefore, the two representations cannot be disentangled at all. 

\subsubsection{Adversarial Asymmetric Training} 
\label{sec:aat}
To further achieve disentanglement, we need to break this symmetry, and the key is nothing but adversarial examples. 
For a data pair $(x,y)$, we can generate an adversarial example $x^{s}_{adv}$ (Eq. \ref{eqn:untargeted-attack-objective}) w.r.t. the standard way loss $l(h_s(x;\theta), y)$ to imitate attack to a standard classifier.
We keep $x^s_{adv}$ if it is misclassified, i.e., its predicted class $\hat y_s\neq y$, and discard it otherwise. Because $x^s_{adv}$ is very close to $x$ (visually indistinguishable), by definition, its robust features are still about label $y$. In contrast, its non-robust features now belong to the misclassified class $\hat y_s$ due to the adversarial attack. 

In this way we obtain an image with robust and non-robust features about different classes. 
This \textbf{asymmetry} enables us to disentangle the two representations with the following bi-level objective, 
\begin{equation}
\label{eqn:adversarial-standard-way-loss}
\begin{aligned}
\cL_{AS}(\theta,x,y) = {\color{darkgreen} l(h_r(x^s_{adv};\theta),y)} + {\color{red}l(h_n(x^s_{adv};\theta),\hat y_s)},\\
\text{s.t.~~}
\begin{cases}
\displaystyle x^s_{adv}=x+\argmax_{\delta\in\cS}\,{\color{black}l(h_s(x+\delta;\theta), y)},\\
\displaystyle \hat y_s=\argmax\, h_s(x^s_{adv};\theta),\,\hat y_s\neq y,
\end{cases}
\end{aligned}
\end{equation}
where we let the robust way predict the original label $y$ from $z_r$ and let the non-robust way predict the perturbed label $\hat y_s$ from $z_n$. Therefore, the two asymmetric labels provide different supervisions to the two representations, such that $z_r$ stays invariant under attack (robust), and $z_n$ becomes sensitive to perturbations (non-robust). Therefore the two representations are disentangled.

The total loss is the combination of standard loss and  adversarial loss for a balance between accuracy and disentanglement, and we re-organize it as
\begin{equation}
\label{eqn:total-aat-loss}
\begin{aligned}
\cL_{total}(\theta,x,y) &=\,\cL_{ST}(\theta,x,y) + \cL_{AS}(\theta,x,y) \\
&=\, {\color{blue} l(h_s(x;\theta),y)} + {\color{darkgreen} l(h_r(x;\theta),y) + l(h_r(x^s_{adv};\theta),y)} \\
&\quad\quad\quad\quad\quad\quad~~
+ {\color{red} l(h_{n}(x;\theta),y) + l(h_n(x^s_{adv};\theta),\hat y_s)},
\end{aligned}
\end{equation}
where $(x^s_{adv},\hat y_s)$ is generated according to Eq. \ref{eqn:adversarial-standard-way-loss}. Overall, with the pseudo-pair $(x,x^s_{adv})$,  we assign different goals to the three ways to learn robust and non-robust representations,
\begin{itemize}
\item \textbf{\color{blue} the standard way  $h_s$} should take both kinds of features to achieve better standard accuracy;
\item \textbf{\color{darkgreen} the robust way $h_r$} is encouraged to be invariant to perturbations as it is supposed to predict label $y$ whether there is an adversary or not;
\item \textbf{\color{red} the non-robust way $h_n$} is taught to be very sensitive to input perturbations, as it predicts $x$ to original class $y$ and predicts $x^s_{adv}$, which is very close to $x$, to a different class $\hat y_s$.
\end{itemize}
To distinguish from Adversarial Training \cite{goodfellow2014explaining} that extracts robust features alone, we call our method \textit{Adversarial Asymmetric Training} (AAT), which instead preserves both robust and non-robust features by disentanglement with asymmetric supervisions. See Figure \ref{fig:diagram} (right) for an example and Algorithm \ref{algo:aat} for a complete description.

As we see, the pseudo-pairs are not given but a result of our training process. Remind that the prior of robustness is human-centric. Thus robustness comes from nowhere but our design of training objectives. In other words, the supervision of robustness is intrinsic rather than extrinsic. In our work, adversarial examples are not only ``not bugs'' \cite{ilyas2019adversarial}, and they instead become the key to the disentanglement of robust and non-robust representations.

\begin{algorithm}[t]
    \textbf{Input:} natural data pair $(x,y)\in\cD_{train}$, current model parameters $\theta$;\\
    \textbf{Output:} training loss $\cL_{total}(\theta,x,y)$;
    
    \begin{algorithmic}
        \STATE Predict $x$ in three ways $h_s, h_r, h_{n}$ (Eq. \ref{eqn:three-ways});
        \STATE Calculate standard loss $\cL_{ST}$ (Eq. \ref{eqn:total-standard-training-loss});
        \STATE Generate an adversarial example $x^s_{adv}$ w.r.t. the standard way loss (Eq. \ref{eqn:untargeted-attack-objective});
        \IF {$x^s_{adv}$ is misclassified} 
            \STATE Predict $x^s_{adv}$ in two ways $h_r, h_n$ (Eq. \ref{eqn:three-ways});
            \STATE Calculate adversarial loss $\cL_{AS} $ (Eq. \ref{eqn:adversarial-standard-way-loss});
            \STATE \textbf{return} loss $\cL_{total}=\cL_{ST}+\cL_{AS}$;
        \ELSE
        \STATE \textbf{return} loss $\cL_{total}=\cL_{ST}$;
        \ENDIF
    \end{algorithmic}
    \caption{A training episode loss computation of Adversarial Asymmetric Training (AAT)}
    \label{algo:aat}
\end{algorithm}

\subsubsection{AAT++}

To further encourage the disentanglement of the two representations, we design two auxiliary asymmetric losses. In particular, to enhance the robustness of $z_r$, we perform robust optimization (Eq. \ref{eqn:adversarial-training}) w.r.t. the robust way $h_r$, which can be written equivalently as
\begin{equation}
\cL_{AR}(\theta,x,y)={\color{darkgreen} l(h_r(x^r_{adv};\theta),\, y)},
\quad \text{ s.t.~~}
x^r_{adv}=x+\argmax_{\delta\in\cS}\,{\color{black}l(h_r(x+\delta;\theta), y)}.
\label{eqn:adversarial-robust-way-loss}
\end{equation}
Similarly, to enhance the non-robustness of $z_n$, we design the following objective for a non-robust-way adversarial example $x^n_{adv}$ misclassified as $\hat y_n$. As discussed above, the non-robust features of $x^n_{adv}$ belong to $\hat y_n$ due to adversarial attack. Therefore, we encourage the non-robust way $h_n$ to detect the non-robust features of $x^n_{adv}$ with supervision $\hat y_n$:
\begin{equation}
    \label{eqn:adversarial-non-robust-way-loss}
\begin{aligned}\quad\quad
    \cL_{AN}(\theta,x,y) = {\color{red} l(h_n(x^n_{adv};\theta),\hat y_n)},
    \quad\text{s.t.~}
    \begin{cases}
    \displaystyle x^n_{adv}=x+\argmax_{\delta\in\cS}\,{\color{black}l(h_n(x+\delta;\theta), y)},\\
    \displaystyle \hat y_n=\argmax\, h_n(x^n_{adv};\theta),\,\hat y_n\neq y.
    \end{cases}
\end{aligned}
\end{equation}
The two auxiliary asymmetric losses here are designed to further ``purify'' each representation to be more robust or non-robust. Hence we coin the name AAT++ with total loss
\begin{equation}
\label{eqn:total-aat-plus-plus-loss}
\begin{aligned}
\cL_{total}^{++}(\theta,x,y) =\,&\cL_{ST}(\theta,x,y)+\cL_{AS}(\theta,x,y) + \cL_{AR}(\theta,x,y) + \cL_{AN}(\theta,x,y) \\
=\,& {\color{blue} l(h_s(x;\theta),y)} + {\color{darkgreen} l(h_r(x;\theta),y) + l(h_r(x^s_{adv};\theta),y) + l(h_r(x^r_{adv};\theta), y)}\\
&~~~~~~~~~~~~~~~~~~~~~
+ {\color{red} l(h_{n}(x;\theta),y) + l(h_n(x^s_{adv};\theta),\hat y_s) + l(h_n(x^n_{adv};\theta), \hat y_n)},
\end{aligned}
\end{equation}
where $(x^s_{adv},\hat y_s), x^r_{adv},(x^n_{adv},\hat y_n)$ are generated according to Eq. \ref{eqn:adversarial-standard-way-loss}, \ref{eqn:adversarial-robust-way-loss} \& \ref{eqn:adversarial-non-robust-way-loss}, respectively. As shown in our ablation study in Sec. \ref{sec:analysis}, the auxiliary terms can enhance the robustness disentanglement in general, at the cost of sacrificing a little standard accuracy.

\subsection{Adversarial Detection}
\label{sec:detection}
Previous works have proposed various heuristics for detecting adversarial examples \cite{xu2017feature,gong2017adversarial,metzen2017detecting, grosse2017statistical,bhagoji2017dimensionality}, yet typically without an understanding of the existence of adversarial examples. Our disentanglement of robust and non-robust features offers a principled approach for adversarial detection.

As discussed previously, the fundamental characteristic of (misclassified) adversarial examples is the disagreement between robust and non-robust features. Therefore, we can detect adversarial examples based on the two disentangled representations. Intuitively, if $z_r$ and $z_n$ agree, it is a natural image, otherwise it is adversarial. Here we give a simplest rule $D(x)$ to illustrate this idea, 
\begin{equation}
\begin{gathered}
    y_r= \argmax h_r(x;\theta),~y_n=\argmax h_n(x;\theta);~~
    D(x)=
    \begin{cases}
        0 \text{ (natural)}, & \text{if }y_r =y_n;\\
        1 \text{ (adversarial)}, &\text{if }y_r \neq y_n,
    \end{cases}
\end{gathered}             
\label{eqn:detection}
\end{equation}
that is, directly comparing the predictions from the two representations. This rule can be directly applied with our three-way model without extra computation. More complex strategies can also be considered to exploit more information from the disentangled representations, e.g. training an additional binary classifier \cite{metzen2017detecting} based on $z_r$ and $z_n$. We leave this for future work.

\subsection{Evaluation Metric}
Based on our model, we propose two evaluation metrics for robustness disentanglement. Similarly, the evaluation also does not require additional supervision on robustness. 

\textbf{Difference in Accuracy (DIA).} For adversarial examples, the robust features are about the original label, while the non-robust features likely belong to a different class. This will lead to the high accuracy of the robust way and low accuracy of the non-robust way. In turn, a larger accuracy gap, namely \emph{Difference in Accuracy}, indicates better disentanglement of the two representations. 

\textbf{Rate of Adversarial Detection (RAD).} We devise a rule for adversarial detection in Sec. \ref{sec:detection} by comparing inferred labels from robust and non-robust representations. Better disentanglement will yield a better detection rate, and in turn, a better detection rate also indicates better disentanglement.

\section{Experiments}
\label{sec:experiments}

\subsection{Setup}
We conduct experiments on two well-known image classification tasks, MNIST and CIFAR-10. More experimental details can be found in Appendix \ref{sec:additional-setup}.

\textbf{Model.} We build our three-way model based on canonical CNNs for image classification. Specifically, we remove the output layer and take the remaining modules as an encoder. We use two such encoders as $g_r$ and $g_n$, and use a multi-layer perceptron on top as $f$. For CIFAR-10, we consider two backbones, WideResNet34 \cite{zagoruyko2016wide} and PreAct-ResNet18 \cite{he2016deep}. For MNIST, we adapt from a small CNN \cite{zhang2019you}. The hyper-parameters are inherited from conventions \cite{zhang2019you} without any additional tuning. 

\begin{table}[t]
\caption{WideResNet34 backbone results on CIFAR-10 (accuracy in percentage).}
\label{tab:wide-cifar-10-comparison}
\begin{tabular}{l|l|ccc|ccc|ccc}
\hline
\multirow{2}{*}{Model} & \multirow{2}{*}{Method} & \multicolumn{3}{c|}{Standard}                  & \multicolumn{3}{c|}{Adversarial ($\ell_\infty$)}  & \multicolumn{3}{c}{Adversarial ($\ell_2$)}         \\ \cline{3-11}
                       &          & S($\uparrow$) & R($\uparrow$) & N($\uparrow$) & R($\uparrow$)   & N($\downarrow$)   & DIA ($\uparrow$)  & R($\uparrow$) & N($\downarrow$) & DIA ($\uparrow$) \\ \hline
\multirow{3}{*}{1-way} & ST (baseline)      & -             & -             & {95.1}          & -               & 0.0               & -                 & -             & 33.8            & -                \\
                       & AT \cite{madry2018towards}      & -             & 90.0          & -             & \textbf{39.9}            & -                 & -                 & \textbf{84.0}          & -               & -                \\
                       & ST+AT & -             & 90.0          & {95.1}          & \textbf{39.9}            & \textbf{0.0}               & \textbf{39.9}              & \textbf{84.0}          & 33.8            & 50.2             \\ \hline
\multirow{3}{*}{3-way} & PI \cite{ilyas2019adversarial}       & 86.8          & 79.9          & 81.9          & 0.0             & \textbf{0.0}               & 0.0               & 39.3          & \textbf{2.3}            & 37.0             \\
                       & AAT (ours)     & \textbf{95.2}          & {91.1}          & 94.7          & 21.7            & \textbf{0.0}               & 21.7              & 81.7          & 25.0            & 56.7             \\
                       & AAT++ (ours)   & 94.1          & 88.7          & 93.7          & \textbf{39.9}            & \textbf{0.0}               & \textbf{39.9}              & 82.5          & 5.3             & \textbf{77.2}             \\ \hline
\end{tabular}
\end{table}
\begin{table}[t]\centering
    \caption{PreAct-ResNet18 backbone results on CIFAR-10 (accuracy in percentage).}
    \label{tab:res-cifar-10-comparison}
\begin{tabular}{l|l|ccc|ccc|ccc}
    \hline
\multirow{2}{*}{Model} & \multirow{2}{*}{Method} & \multicolumn{3}{c|}{Standard}                  & \multicolumn{3}{c|}{Adversarial ($\ell_\infty$)}  & \multicolumn{3}{c}{Adversarial ($\ell_2$)}         \\ \cline{3-11}
                       &                           & S($\uparrow$) & R($\uparrow$) & N($\uparrow$) & R($\uparrow$) & N($\downarrow$) & DIA ($\uparrow$) & R($\uparrow$) & N($\downarrow$) & DIA ($\uparrow$) \\ \hline
\multirow{3}{*}{1-way} & ST (baseline)                       & -             & -          & {94.2}             & -                & \textbf{0.0}                & -                & -                & 35.1               & -                \\
                       & AT \cite{madry2018towards}                       & -             & 89.0             & -          & \textbf{35.7}             & -                  & -                & \textbf{81.5}             & -                  & -                \\
                       & ST+AT                  & -             &  89.0         & {94.2}          & \textbf{35.7}             & \textbf{0.0}                & \textbf{35.7}             & \textbf{81.5}             & 35.1               & 46.4             \\ \hline
\multirow{3}{*}{3-way} & PI \cite{ilyas2019adversarial}              & 86.8          & 79.9          & 81.9          & 0.0              & 0.0                & 0.0              & 41.4             & \textbf{0.8}                & 40.6             \\
                       & AAT (ours)                      & \textbf{94.8}          & {91.8}          & 93.8          & 10.8             & 0.0                & 10.8             & 79.9             & 22.1               & \textbf{57.8}             \\
                       & AAT++ (ours)                    & 94.2          & 88.2          & 93.7          & 33.5             & \textbf{0.0}                & 33.5             & 81.0             & 28.2               & 52.8             \\ \hline
\end{tabular}
\end{table}

\begin{table}[t]\centering
\caption{MNIST classification results (accuracy in percentage).}
\label{tab:mnist-comparison}
\begin{tabular}{l|l|ccc|ccc}
\hline
\multirow{2}{*}{Model} & \multirow{2}{*}{Method} & \multicolumn{3}{c|}{Standard}                  & \multicolumn{3}{c}{Adversarial ($\ell_2$)}                    \\ \cline{3-8}
                       &                           & S($\uparrow$) & R($\uparrow$) & N($\uparrow$) & R($\uparrow$) & N($\downarrow$) & DIA ($\uparrow$) \\ \hline
\multirow{3}{*}{1-way} & ST (baseline)                       & -             & -             & 99.5          & -             & 17.9            & -                \\
                       & AT \cite{madry2018towards}                       & -             & 99.5          & -             & \textbf{90.1}          & -               & -                \\
                       & ST+AT                     & -             & 99.5          & 99.5          & \textbf{90.1}          & 17.9            & 72.2             \\ \hline
\multirow{2}{*}{3-way} & AAT (ours)                      & \textbf{99.6}          & 99.6          & 99.5          & 82.9          & 4.6             & 78.3             \\ 
                       & AAT++ (ours)                    & \textbf{99.6}          & 99.5          & 99.4          & 89.9          & \textbf{0.0}             & \textbf{89.9}             \\ \hline
\end{tabular}
\end{table}

\begin{table}[t]\centering
\caption{(a) Configurations for adversarial attack, with range $\varepsilon$, $\ell_2$ or $\ell_\infty$ norm, step size $\alpha$, and  $k$ steps of PGD \cite{madry2018towards}. (b) Adversarial detection and calibration results (accuracy in percentage). The model is evaluated on a equal mixture of natural and (standard-way) adversarial examples of CIFAR-10. 
RAD: rate of adversarial detection. Raw/Calibrated: classification accuracy before/after the calibration.}

\begin{subtable}{0.58\linewidth}
    \caption{Adversarial configurations.}
    \begin{tabular}{l|l|cccc}
    \hline
    Data                      & Mode  & Norm          & $\varepsilon$ & $\alpha$ & $k$  \\ \hline
    \multirow{3}{*}{CIFAR-10} & Train & $\ell_\infty$ & 8/255         & 2/255    & 10 \\ \cline{2-6}
                              & \multirow{2}{*}{Test}  & $\ell_\infty$ & 8/255         & 2/255    & 20 \\
                              &   & $\ell_2$      & 0.3           & 0.1      & 20 \\ \hline
    \multirow{2}{*}{MNIST}    & Train & $\ell_2$      & 0.3           & 0.01      & 5 \\ \cline{2-6}
                              & Test  & $\ell_2$      & 0.3           & 0.01      & 10 \\ \hline
    \end{tabular}
    \label{tab:attack-config}
\end{subtable}
\begin{subtable}{0.4\linewidth}\centering
    \caption{Adversarial Detection.}
    \label{tab:detection}
    \begin{tabular}{l|c|cc}
    \hline
    Method    & RAD  & Raw  & Calibrated  \\
    \hline
    PI \cite{ilyas2019adversarial} & 4.9  & 4.6  & 15.2 \\
    AAT          & 64.8 & 61.8 & 65.5 \\
    AAT++        & \textbf{68.8} & \textbf{67.1} & \textbf{69.1} \\
    \hline
    \end{tabular}
\end{subtable}
\end{table}

\textbf{Evaluation.} 
In the test stage, we evaluate natural images via all three ways, denoted as S (standard), R (robust), N (non-robust). 
For white-box adversarial attack (see Table \ref{tab:attack-config} for details), we evaluate the robust way with adversarial examples generated w.r.t. the robust-way loss, and likewise for the non-robust way. For completeness, we also include results for adversarial attack via the standard way, and our methods produce more promising results in this scenario. See Appendix \ref{sec:standard-way-attack-results}. 

We also implement previous methods within our three-way model for a fair comparison. Note our implementations achieve comparable performance to the original work \cite{tsipras2018robustness,zhang2019you}.
\begin{itemize}
\item \textbf{One-way ST \& AT.} Because the robust and non-robust ways utilize only one encoder and one representation, they are almost identical to a normal CNN classifier. Thus we use a single way of our model to implement traditional methods: Standard Training (ST) with the non-robust way $h_n$, and Adversarial Training (AT) \cite{madry2018towards} with the robust way $h_r$.\footnote{We compare with half-half Adversarial Training \cite{tsipras2018robustness} for a balance between standard and robust accuracy.}  
\item \textbf{ST+AT.} In fact, the simplest solution to extract robust and non-robust representations would be to combine a standard and robust classifier, though they may not be properly aligned. We can easily evaluate this method by combining the results of two separate models with one-way ST and AT as above.
\item \textbf{PI (Pseudo-Input)} \cite{ilyas2019adversarial}. The authors offer robust and non-robust versions of CIFAR-10 as pseudo inputs.\footnote{\url{https://github.com/MadryLab/constructed-datasets}} We use them to learn disentangled representations in our three-way model. We train the robust way with the robust dataset and train the non-robust way with the non-robust datasets.
\end{itemize}

\subsection{Classification Results}
The quantitative results are illustrated in Table \ref{tab:wide-cifar-10-comparison}, \ref{tab:res-cifar-10-comparison} \& \ref{tab:mnist-comparison}. 
We mainly take WideResNet34 and $\ell_2$ attack in Table \ref{tab:wide-cifar-10-comparison} for discussion and the rest are similar.  

\textbf{Standard accuracy.} Comparing the standard accuracy of one-way ST and AT, we can see that adversarial training leads to a severe accuracy reduction ($\sim$5\%) as it discards non-robust features. Instead, in our three-way model with AAT, the standard way effectively improves standard accuracy and is even competitive with standard training. AAT++ leads to a slight accuracy drop, but only by one percent.
It shows our method successfully preserves standard accuracy by combining robust and non-robust features. However, another disentanglement approach, the pseudo-input method \cite{ilyas2019adversarial}, leads to even worse standard accuracy (86.8\%), indicating its iterative optimization over the input causes severe loss of details in the raw image. 

\textbf{Adversarial accuracy and disentanglement.} The pseudo-input method indeed achieves some degree of disentanglement, which is, however, very limited (37.0\% DIA), and diminishes quickly under stronger attack ($\ell_\infty$). In comparison, our pseudo-pair methods achieve much better disentanglement with better robust accuracy, 56.7\% DIA of AAT and 77.2\% DIA of AAT++. 
As for one-way methods, AT has the best robust accuracy (84.0\%), yet the robust way of our AAT++ nearly matches this limit (82.5\%) and is also competitive under stronger attack. Meanwhile, the non-robust way of AAT++ achieves much lower accuracy than one-way ST (5.3\% v.s. 33.8\%), indicating better non-robustness. Consequently, the disentanglement of our three-way model is better than the combination of two one-way models (77.2\% v.s. 50.2\% DIA). The advantage is more evident when the attack is relatively weak.

\subsection{Further Analysis}
\label{sec:analysis}

\begin{figure}[t]\centering
\centering
\begin{subfigure}{.4\textwidth}\centering
    \centering
    \includegraphics[width=\linewidth]{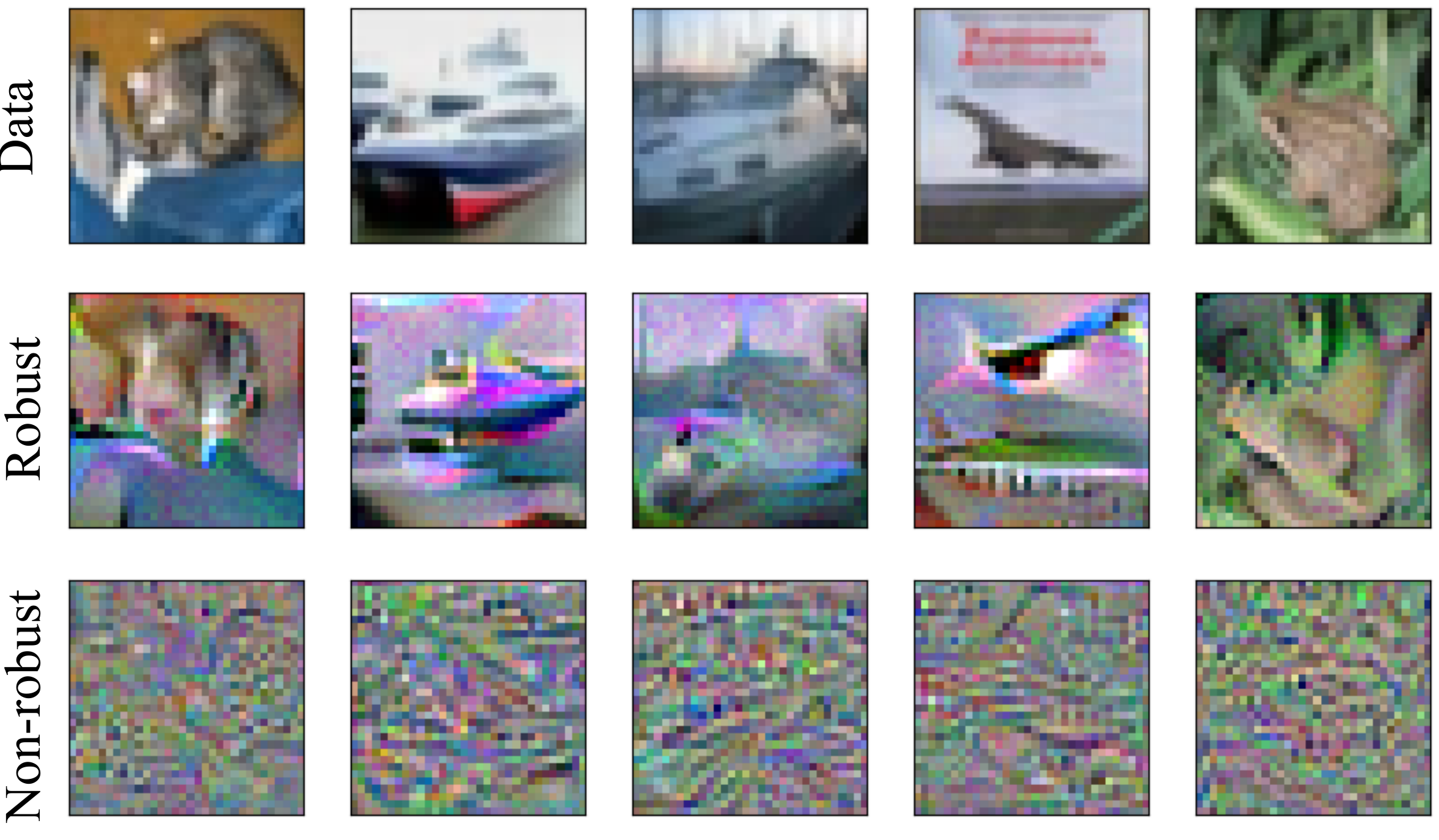}
    \caption{Representation Inversion.}
    \label{fig:inversion}
\end{subfigure}
\begin{subfigure}{.57\textwidth}\centering
\includegraphics[width=\linewidth]{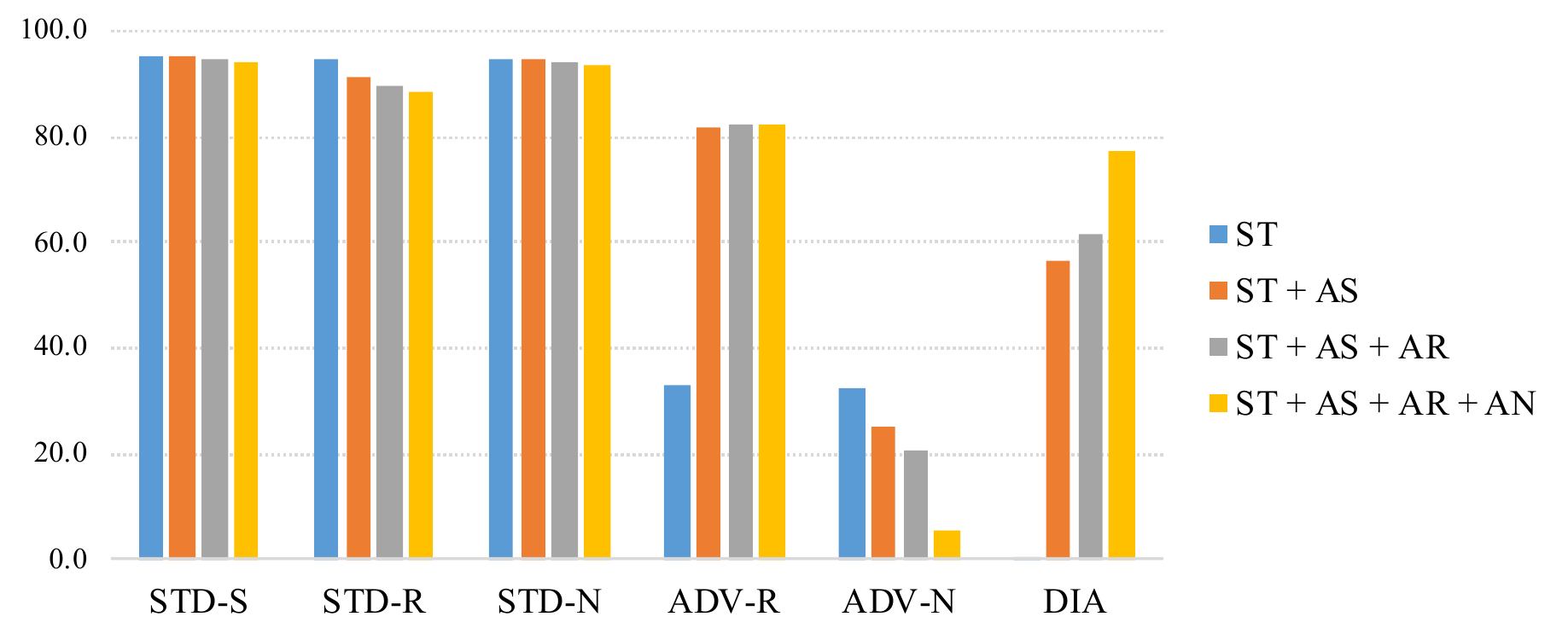}
\caption{Ablation study.}
\label{fig:ablation-study}
\end{subfigure}
\caption{(a) Visualization of robust and non-robust representations learned by AAT++. (b) Ablation study of the four losses of AAT++ (Eq. \ref{eqn:total-aat-plus-plus-loss}), with WideResNet34 backbone and $\ell_2$ attack on CIFAR-10. STD: standard. ADV: adversarial.}
\label{fig:figures}
\end{figure}

\textbf{Ablation study.} We conduct ablation study for the four losses of AAT++ (Eq. \ref{eqn:total-aat-plus-plus-loss}), as shown in Figure \ref{fig:ablation-study}. $\cL_{ST}$ achieves good standard accuracy but yields no disentanglement. The introduction of $\cL_{AS}$ effectively achieves the disentanglement of two representations. Furthermore, adding $\cL_{AR}$ improves the robust accuracy, and adding $\cL_{AN}$  brings down the non-robust accuracy significantly. Combining four terms as in AAT++, we have the best disentanglement with the highest DIA score, while the standard accuracy drops a little in the meantime.

\textbf{Adversarial detection and calibration.} We evaluate our na\"ive detection rule (Eq. \ref{eqn:detection}) and results are shown in Table \ref{tab:detection}. The pseudo-input method performs much worse than random guess (4.9\% RAD) because its robust accuracy is too poor. This also suggests its disentanglement is very limited. AAT and AAT++ instead enjoy considerably better detection rates. More complex strategies can be utilized for further improvement. As an additional application, we can also use the detection rule to calibrate our prediction for the mixture of natural and adversarial images. Specifically, we apply the robust way for inferred adversarial images and apply the standard way otherwise. From Table \ref{tab:detection}, we can see that the calibration helps improve classification accuracy in total.

\textbf{Visualization.}
To intuitively understand the disentanglement, we invert the two representations to input-level following \cite{engstrom2019learning}. From Figure \ref{fig:inversion}, we can see that the inversion of the robust representation is human-conceivable, while that of non-robust representation seems just plain noise. 
This is consistent with the phenomenon in previous work \cite{engstrom2019learning} that robust features are perceptually aligned with humans, while non-robust features are not. More qualitative results are included in Appendix \ref{sec:additional-qualitative-results}.

\section{Related Work}

Adversarial examples are proposed as a threat to machine learning models \cite{szegedy2013intriguing}. Afterwards, Adversarial Training \cite{goodfellow2014explaining} is developed to enhance robustness by feeding adversarial examples while training. Although effective, AT is found to be the cause of severe accuracy reduction, and the trade-off between accuracy and robustness is fundamentally inevitable \cite{tsipras2018robustness,zhang2019theoretically}. Nevertheless, adversarial examples are not thus put to death and become useless. Recently, it is shown that it is possible to improve standard accuracy with adversarial examples \cite{xie2019adversarial}. Our work also contributes to this thread as we find adversarial examples can also serve as the fuel for disentangled representation learning.

Nevertheless, our method is not the only approach to utilize adversarial methods for disentanglement. AdvMix \cite{gowal2019achieving} instead disentangles ``relevant'' and ``irrelevant'' features w.r.t. class with a minimax game, while our work focuses on robustness disentanglement. However, AdvMix relies crucially on a pre-trained StyleGAN \cite{karras2019style}, while our method trains from scratch and is decoder-free.

In semi-supervised learning scenarios, previous works utilize virtual labels, i.e., the current inferred labels, to conduct adversarial training for unsupervised data \cite{miyato2018virtual,stanforth2019labels,zhang2019theoretically}. In this work, we instead use virtual labels, $\hat y_s$ and $\hat y_n$, as supervisions for the non-robust features of adversarial examples. The virtual labels are found to work well as long as the current model is relatively precise. 

\section{Conclusion}
In this paper, we have developed a novel Adversarial Asymmetric Training scheme for disentangling robust and non-robust representations without additional supervision on robustness. Our method is decoder-free, end-to-end, and achieves much better disentanglement compared to previous methods. Future work may include more efficient architecture designs, applications to other computer vision tasks, as well as advanced adversarial detection methods based on the disentangled representations. 

\bibliographystyle{plain}
\bibliography{robust}


\appendix

\section{Additional Theoretical Discussions}

\subsection{Definitions of Robust and Non-Robust Representations}
\label{sec:representation-definitions}
Similar to the definitions of robust and non-robust features delivered in Sec. \ref{sec:preliminary}, here we present our definitions for robust and non-robust representations accordingly.

Define a representation $r$ as a function mapping from the input space $\cX$ to a latent Space $\cZ$ of lower dimension, $r:\cX\to\cZ$. We further define a classifier $f$ as a function mapping from the latent Space $\cZ$ to real numbers, $f:\cZ\to\bbR$. 
With a specified data distribution $\cD$ and an adversarial configuration $\cS$, we give formal definitions of robust and non-robust representations for binary classification ($C=2$).
\begin{itemize}
    \item 
    We call a representation $r$ \emph{$\rho$-useful} ($\rho$ > 0) if there exists a classifier $f$, such that $h=f\circ r$ is correlated with the true label in expectation, i.e.,
$        \mathbb{E}_{(x, y) \sim \mathcal{D}}[y \cdot h(x)] \geq \rho.$
    \item 
    Suppose we have a \(\rho\)-useful representation $r$, we refer to \(r\) as a \emph{robust representation} (formally a $\gamma$-robustly useful representation) if, there exists a classifier $f$, such that $h=f\circ r$ remains \(\gamma\)-useful $(\gamma>0)$ under adversarial perturbation, i.e.,
    $ \mathbb{E}_{(x, y) \sim \mathcal{D}}\left[\min _{\delta \in\cS} y \cdot h(x+\delta)\right] \geq \gamma.$
    \item We refer to \(r\) as a \emph{non-robust representation} (formally a $\gamma$-non-robustly useful representation) 
     if it is $\rho$-useful for some $\rho>0$, but not $\gamma$-robust ($\gamma> 0$).
\end{itemize}

\subsection{The Accuracy-Robustness Dilemma}
\label{sec:dilemma}
As discussed in the main text, there is a fundamental trade-off between accuracy and robustness that enhancing robustness will inevitably lead to the degradation of standard accuracy. Besides the empirical evidence given in previous work \cite{tsipras2018robustness}, it is shown that the accuracy-robustness dilemma exits even if we have infinite data and optimal classifiers. Here, we present such an example to illustrate the phenomenon. 
Our example is a variation of the one presented in \cite{tsipras2018robustness}, which we review as follows.

\textbf{The binary classification problem in \cite{tsipras2018robustness}.} The data model consists of input-label pairs $(x, y)$ sampled from a distribution $\mathcal{D}$ as follows:
\begin{equation}
    y \stackrel{u . a \cdot r}{\sim\{-1,+1\},} \quad x_{1}=\left\{\begin{array}{ll}
        +y, & \text { w.p. } p \\
        -y, & \text { w.p. } 1-p^{\prime}
        \end{array} \quad x_{2}, \ldots, x_{d+1} \stackrel{i . i . d}{\sim} \mathcal{N}(\eta y, 1)\right.
\end{equation}
For this problem, a natural classifier 
\begin{equation}
    f_{\operatorname{avg}}(x):=\operatorname{sign}\left(w_{\text {unif }}^{\top} x\right), \quad \text { where } w_{\text {unif }}:=\left[0, \frac{1}{d}, \dots, \frac{1}{d}\right]
\end{equation}
achieves standard accuracy arbitrarily close to $100$\%,  for $d$ large enough. 
However, an $\ell_\infty$-bounded adversary with $\varepsilon=2\eta$, can shift the weakly-correlated features $\{x_2,\dots,x_{d+1}\}$ towards $-y$. As a result, the simple classifier that relies solely on these non-robust features cannot get adversarial accuracy better than $1\%$.

On the contrary, assume $p>0.5$, the classifier relying solely on robust features
\begin{equation}
f_{rob}(x)=\operatorname{sign}\left(w_{rob}^{\top}x\right),\quad\text{ where }w_{rob}:=[1,0,\dots,0]
\end{equation}
will have $p$ accuracy in expectation under both standard and adversarial scenarios. $f_{rob}$ attains better robustness, at the cost of sacrificing accuracy ($p<1$). Such a trade-off between accuracy and robustness is fundamental and will not disappear even with infinite samples and Bayes-optimal classifiers. 

\textbf{Limitations.} The example above gives a clear illustration of the dilemma. However, the problem setup is somewhat misleading because it suggests that a classifier relying solely on non-robust features can achieve optimal accuracy. In fact, as shown in our experiments, such a classifier usually has sub-optimal standard accuracy, similar to the robust classifier. Only the standard way that combines robust and non-robust features can achieve comparable performance to the one-way ST model. Here, we give a variation of their example that shows not only the accuracy-robustness dilemma but also the benefit of preserving both robust and non-robust features.

\textbf{Our example.} We consider the case where input pair \((x,y)\) follows distribution
\begin{gather}
y \stackrel{u, a, r}{\sim}\{-1,+1\}, x_{i}=\left\{\begin{array}{ll}{+\eta_i y,} & {\text {w.p. } p} \\ {-\eta_iy,} & {\text {w.p. } 1-p}\end{array},\right. i\in[d].
\end{gather}
For simplicity we assume $p=0.8, d=7,$ and
\begin{equation*}
\eta_1=\eta_2=\eta_3=\eta_4=0.01,~\eta_5=\eta_6=\eta_7=1.
\end{equation*}

\begin{prop}
For the problem above, we have the following conclusions:
\begin{enumerate}[label=\arabic*)]
\item The following linear classifier 
\begin{gather}
    h_0(x):=\operatorname{sign}\left(w^{\top}_0 x\right),
    w_0=\left[\frac{1}{\eta_1},\frac{1}{\eta_2},\dots,\frac{1}{\eta_7}\right]^\top.
\end{gather} has $94.4\%$ standard accuracy in expectation and it is Bayes-optimal. Nevertheless, it always has 0\% accuracy under $\ell_\infty$ attack with $\varepsilon=0.02$, because the adversary can shift the non-robust features $[x_1,x_2,x_3,x_4]$, towards $-y$.
\item Instead, the classifier
\begin{equation}
    h_1(x):=\operatorname{sign}\left( w^{\top}_1 x\right), w_1=\left[0,0,0,0,\frac{1}{\eta_5},\frac{1}{\eta_6},\frac{1}{\eta_7}\right]^\top.
\end{equation}
is optimal under attack, because it relies solely on the last three robust features $[x_5,x_6,x_7]$. $h_1$ attains 88.0\% accuracy in expectation for both standard and adversarial scenarios. 
\end{enumerate}
\end{prop}

\textbf{Discussion.} In our problem setup, neither robust nor non-robust features are perfect. Each of them can be misleading for the correct label with a certain probability, and their combination can average out the risks and yield optimal standard accuracy, which is more consistent with our experimental results discussed in Sec. \ref{sec:experiments}. It also suggests that standard training that aims at best accuracy is not enough to extract solely non-robust features. Instead, besides the usefulness pursued by standard training, our AAT and AAT++ further enhance the non-robustness of our non-robust way by enforcing it to be sensitive to adversarial perturbations.

\begin{proof}
1) For simplicity, we denote $\hat x_i=x_i/\eta_i$ for re-weighted features. Then the linear classifier
\[h_0(x)=\operatorname{sign}\left(\sum_{i=1}^7\hat x_i\right),\quad\hat x_i\in\{\pm 1\},\]
is equivalent to a majority voting method, and its classification is wrong only when there are at least four $x_i$'s indicating $-y$. Hence the expected standard accuracy of $h_0$ follows
\begin{equation}
\label{eqn:standard-acc}
\begin{aligned}
P\left(h_0(x)=y\right)=1-C_{7}^{4}(1-p)^4=1-35\times 0.0016=0.944.
\end{aligned}
\end{equation}
Next, we prove its optimality. According to the problem setup, we have
\begin{equation}
\begin{aligned}
P(Y=y|X=x)=&\frac{P(X=x|Y=y)P(Y=y)}{P(X=x)}=\prod_i\frac{p^{(\hat x_iy+1)/2}(1-p)^{(1-\hat x_iy)/2}}{\sum_{y} p^{(\hat x_iy+1)/2}(1-p)^{(1-\hat x_iy)/2}}.
\end{aligned}
\end{equation}
And the decision rule of the Bayes-optimal classifier should be
\begin{equation}
h^*(x)=
\begin{cases}
 +1, & \text{if }P(Y=y|X=x)\geq\frac{1}{2};\\
 -1, & \text{otherwise}.
\end{cases}
\end{equation}
Notice that if there are 4 $\hat x_i$'s different from $y$, we have
\begin{equation}
\begin{aligned}
P(Y=y|X=x)=\frac{p^3(1-p)^4}{p^3(1-p)^4+p^4(1-p)^3}\approx 0.2<0.5,
\end{aligned}
\end{equation}
and the probability is smaller with more such features. So the Bayes-optimal classifier is right when they are at most $3$ $\hat x_i$'s different from $y$. Following Eq. \ref{eqn:standard-acc}, we can conclude that the expected standard accuracy of the Bayes-optimal classifier could be no more than $0.944$. Because $h_0$ achieves this expected accuracy, it is optimal.

2) Because the specified adversary can change the sign of non-robust features arbitrarily, the first four features $[x_1,x_2,x_3,x_4]$ becomes non-informative. Hence we can only rely on the robust features $[x_5,x_6,x_7]$. Following the same deduction in Eq. \ref{eqn:standard-acc}, $h_1$'s expected robust accuracy  is
\begin{equation}
\begin{aligned}
P\left(h_1(x)=y\right)&=1-C_{3}^{2}(1-p)^2=1-3\times 0.04=0.88.
\end{aligned}
\end{equation}
In fact, it is optimal under adversarial attack. It is easy  to tell that $h_1$'s expected standard accuracy of  is also 88\%. Compared to $h_0$, it becomes much more robust, but at the cost of sacrificing $6.4\%$ standard accuracy. $h_0$ can achieve the best standard accuracy because it preserves both robust and non-robust features in the input. It shows that the trade-off between accuracy and robustness occurs even with infinite data and Bayes-optimal classifiers.
\end{proof}

\section{Additional Experimental Setup}
\label{sec:additional-setup}
\subsection{Model} 
In our three-way model, the two encoders, $g_r$ and $g_n$, are beheaded classification models with the same architecture. Hence they provide two representations of the same size $H$. As for the shared classifier on top, the first linear layer has a size of $2H\times H$, followed by a ReLU activation and another linear layer of the size $H\times C$, where $C$ is the number of classes.

\subsection{Training} 
We list our training configurations in Table \ref{tab:training-configurations}. All these hyper-parameters are directly immigrated from the YOPO repository, see \url{https://github.com/a1600012888/YOPO-You-Only-Propagate-Once}. We implement our methods with PyTorch and conduct experiments on NVIDIA P100 GPUs.

\begin{table*}[h]\centering
    \caption{Training Configurations in our experiments.}
    \label{tab:training-configurations}
    \begin{tabular}{l|cccccc}
    \hline
    Task                      & Backbone  & Learning Rate & Momentum & Weight Decay & Epoch & Milestones \\ \hline
    \multirow{2}{*}{CIFAR-10} & Wide34 & 0.1 & 0.9 & 2e-4 & 105  & [75, 90,100] \\ 
                              & Res18 & 0.05 & 0.9 & 5e-4 & 105 & [75, 90,100] \\ \hline
    {MNIST}    & CNN & 0.1 & 0.9 &  5e-4 & 56 & [50,55] \\  \hline
    \end{tabular}
\end{table*}

\section{Additional Quantitative Results}
\label{sec:standard-way-attack-results}
In Sec. \ref{sec:experiments}, we show adversarial accuracies obtained under attack w.r.t. the robust and non-robust way loss, respectively. 
In this part, we include additional results when the attack is crafted w.r.t. the standard way of our model, which resembles the attack to a standard classifier. 

From Table \ref{tab:cifar10-wide}, \ref{tab:cifar10-res}, \ref{tab:mnist-standard}, we can see AAT can yield even better robust accuracy than one-way AT under standard-way attack, and AAT++ can generally achieve better robustness. The disentanglement score is also better than the pseudo-input method. 

Nevertheless, we notice that our non-robust accuracy is not as good as the pseudo-input method. Meanwhile, its non-robust-way standard accuracy is much lower than ours, due to the loss of details in the raw image. It indicates that the combination of the pseudo-input and pseudo-pair methods might produce better disentanglement without loss of accuracy. We leave this for future work.

\begin{table}[h]\centering
    \caption{CIFAR-10 results (accuracy in percentage) with WideResNet34 backbone  with attack w.r.t. the standard way.}
    \label{tab:cifar10-wide}
    \begin{tabular}{l|l|cccc|cccc}
    \hline
    \multirow{2}{*}{Model} & \multirow{2}{*}{Training} & \multicolumn{4}{c|}{Adversarial   ($\ell_\infty$)}   & \multicolumn{4}{c}{Adversarial   ($\ell_2$)}          \\ \cline{3-10}
                           &                           & S(-) & R($\uparrow$) & N($\downarrow$) & DIA ($\uparrow$) & S(-) & R($\uparrow$) & N($\downarrow$) & DIA ($\uparrow$) \\ \hline
    \multirow{2}{*}{1-way} & ST                        & -    & -             & 0.0             & -                & -    & -             & 33.8            & -                \\
                           & AT \cite{madry2018towards}                       & -    & 39.9          & -               & -                & -    & 84.0          & -               & -                \\ \hline
    \multirow{3}{*}{3-way} & PI \cite{ilyas2019adversarial}                        & 0.0  & 5.2           & 0.0             & 5.2              & 22.5 & 64.8          & \textbf{4.9}             & 59.9             \\
                           & AAT (ours)                      & 0.0  & 59.5          & 0.0             & 59.5             & 69.0 & 89.8          & 35.2            & 54.6             \\
                           & AAT++ (ours)                    & 9.8  & \textbf{66.4}          & 0.0             & \textbf{66.4}             & 77.9 & \textbf{89.0}          & 23.7            & \textbf{65.3}             \\ \hline
    \end{tabular}
\end{table}

\begin{table}[h]\centering
    \caption{CIFAR-10 results (accuracy in percentage) with Pre-activated ResNet18 backbone  with attack w.r.t. the standard way.}
    \label{tab:cifar10-res}
    \begin{tabular}{l|l|cccc|cccc}
    \hline
    \multirow{2}{*}{Model} & \multirow{2}{*}{Training} & \multicolumn{4}{c|}{Adversarial   ($\ell_\infty$)}         & \multicolumn{4}{c}{Adversarial   ($\ell_2$)}              \\ \cline{3-10}
                           &                           & S(-) & R($\uparrow$) & N($\downarrow$) & DIA ($\uparrow$) & S(-) & R($\uparrow$) & N($\downarrow$) & DIA ($\uparrow$) \\ \hline
    \multirow{2}{*}{1-way} & ST                        & -    & -             & 0.0             & -                & -    & -             & 35.1            & -                \\
                           & AT \cite{madry2018towards}                       & -    & 35.7          & -               & -                & -    & 81.5          & -               & -                \\ \hline
    \multirow{3}{*}{3-way} & PI \cite{ilyas2019adversarial}                     & 0.0  & 8.2           & 0.0             & 8.2              & 21.6 & 65.7          & \textbf{1.3}             & \textbf{64.4}             \\
                           & AAT (ours)                      & 0.2  & 38.9          & 0.0             & 38.9             & 66.6 & 88.1          & 31.4            & 56.7             \\
                           & AAT++ (ours)                    & 1.3  & \textbf{63.8}          & 0.0             & \textbf{63.8}             & 68.4 & \textbf{86.5}          & 35.3            & 51.2             \\ \hline
    \end{tabular}
\end{table}

\begin{table}[h]\centering
    \caption{MNIST results (accuracy in percentage) with attack w.r.t. the standard way.}
    \label{tab:mnist-standard}
    \begin{tabular}{l|l|cccc}
    \hline
    \multirow{2}{*}{Model} & \multirow{2}{*}{Training} & \multicolumn{4}{c}{Adversarial   ($\ell_2$)}              \\ \cline{3-6}
                           &                           & S(-) & R($\uparrow$) & N($\downarrow$) & DIA ($\uparrow$) \\ \hline
    \multirow{2}{*}{1-way} & ST                        & -    & -             & 17.4            & -                \\
                           & AT \cite{madry2018towards}                       & -    & 89.8          & -               & -                \\ \hline
    \multirow{2}{*}{3-way} & AAT (ours)                      & 51.0 & 93.9          & 30.4            & 63.5             \\
                           & AAT++ (ours)                    & 12.6 & 97.8          & 7.3             & 90.5             \\ \hline
    \end{tabular}
\end{table}

\section{Additional Qualitative Results}
\label{sec:additional-qualitative-results}
\begin{figure}[h]
    \centering
    \includegraphics[width=\linewidth]{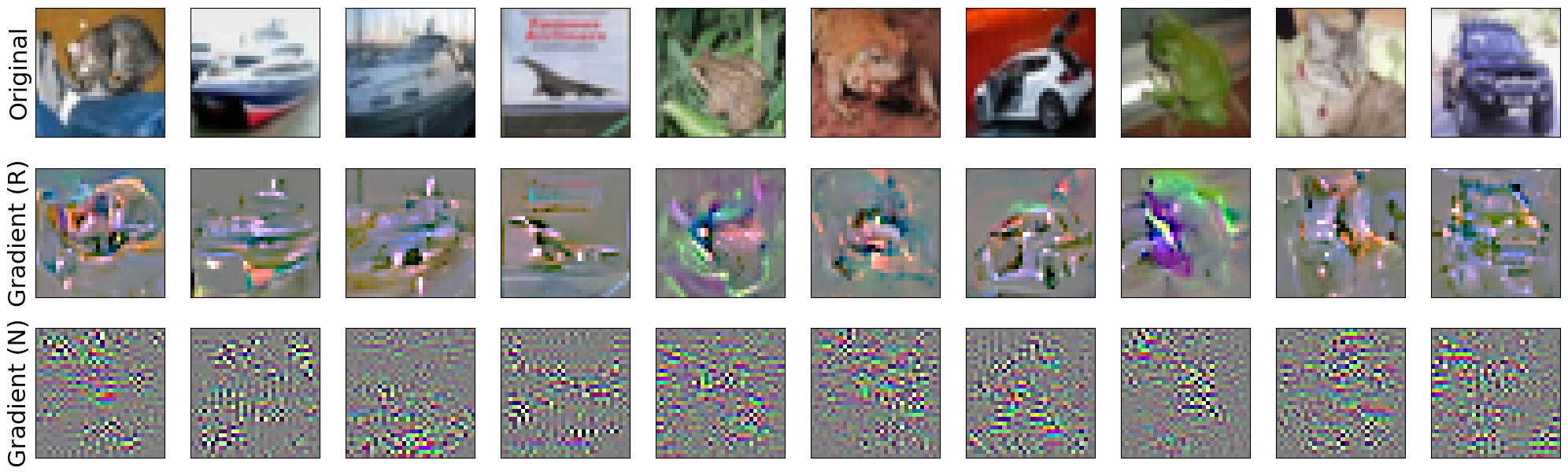}
    \caption{Gradient visualization with WideResNet34 backbone on CIFAR-10.}
    \label{fig:cifar10-grad}
\end{figure}

\begin{figure}[h]
    \centering
    \includegraphics[width=\linewidth]{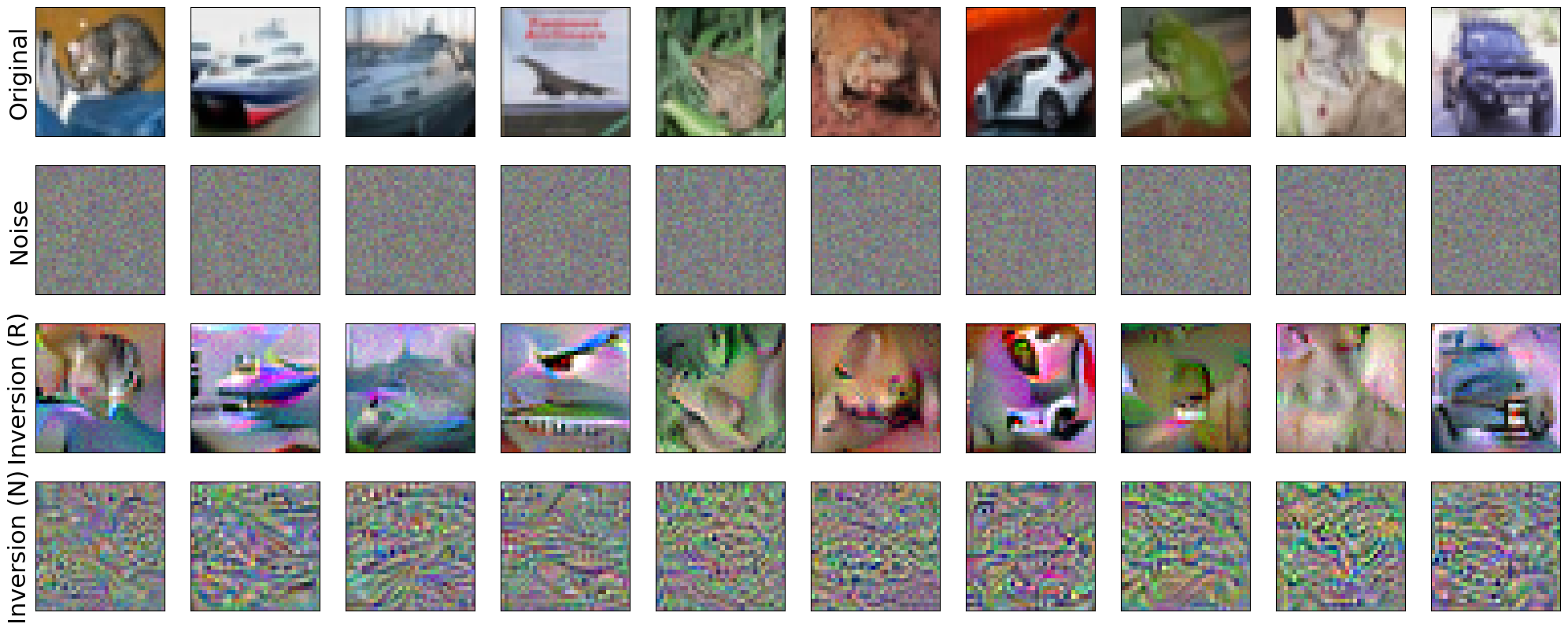}
    \caption{Representation inversion with WideResNet34 backbone on CIFAR-10.}
    \label{fig:cifar10-rep}
\end{figure}

Here, in Figure \ref{fig:cifar10-grad} \& \ref{fig:cifar10-rep}, we give more qualitative results to show the difference between two disentangled representations, including the gradient visualization task \cite{tsipras2018robustness} and the representation inversion task \cite{engstrom2019learning}. They both show that the robust representations are perceptually aligned with humans while the non-robust representations seem plain noise. Details of implementation are as follows.

\subsection{Gradient Visualization}

We show the gradient for a clean image w.r.t. the robust and non-robust way loss as in \cite{tsipras2018robustness}. For each image pair $(x,y)$ and the specified classifier $h$, we calculate the gradient w.r.t. the input $\nabla_x\,l(h(x;\theta),y)$. We clip gradients to $\pm 3$ standard deviations of their mean and rescale them to the range $[0,1]$.

\subsection{Representation Inversion} 

Following \cite{engstrom2019learning}, we inverse each representation to input and see what we can get. We initialize the input from random noise and optimize it by minimizing the distance between its robust (or non-robust) representation and the target representation, similar to the construction of robust and non-robust datasets in \cite{ilyas2019adversarial}. We use a learning rate of 1.0 with 10000 steps for CIFAR-10.

\end{document}